\documentclass[11pt,a4paper]{article}
\usepackage[hyperref]{emnlp-ijcnlp-2019}
\usepackage{times}
\usepackage{latexsym}
\usepackage{graphicx}
\usepackage{multirow}

\usepackage{url}

\aclfinalcopy 

\title{Fast End-to-End Wikification}

\author{Ilya Shnayderman, Liat Ein-Dor, Yosi Mass, Alon Halfon, Benjamin Sznajder,\\
\textbf{
Artem Spector, Yoav Katz, Dafna Sheinwald, Ranit Aharonov and Noam Slonim}  \\
  IBM Research AI\\
  Haifa University, Mount Carmel, Haifa, HA 31905, Israel\\
  {\tt \{ilyashn,liate,yosimass,alonha,benjams,artems,katz,dafna,ranita,noams\}}@il.ibm.com}

\date{}

\begin{document}
\maketitle
\begin{abstract}
  Wikification of large corpora is beneficial for various NLP applications. 
  Existing methods focus on quality performance rather than run-time, and are therefore non-feasible for large data. Here, we introduce RedW, a run-time oriented Wikification solution, 
  based on Wikipedia redirects, that can Wikify massive corpora with competitive performance.
  We further propose an efficient method for estimating RedW confidence, opening the door for applying more demanding methods only on top of RedW lower-confidence results. 
  Our experimental results support the validity of the proposed approach.
\end{abstract}
\section{Introduction}
\label{sec:introduction}
The two key challenges faced by end-to-end (e2e) Wikification systems are \textit{Spotting}, i.e. identifying the terms in a text\footnote{A term can be one or more tokens} that should be Wikified, and \textit{Entity Linking (EL) to Wikipedia}, identifying the relevant Wikipedia page among a set of candidate pages. The latter task is usually more runtime demanding, due to the need to consider the context of the mention when aiming to select the most relevant page. 

Existing e2e Wikification methods \cite{TAGME, WAT,wikify} are 
focused on achieving maximal precision and/or recall, while run time is considered a secondary issue. 
However, many NLP tasks, such as argumentation mining \citep{W17-5110}
and question answering \citep{conf/emnlp/BerantCFL13} 
can 
benefit from Wikifying large corpora, marking the need for light-weight solution.



Here, we present RedW\footnote{A demo is available at \href{http://bit.ly/redwe2e}{http://bit.ly/redwe2e}}, a highly efficient e2e context-free wikification layer
with competitive precision, 
that relies on the wisdom of the crowd as reflected through ``redirect'' pages of Wikipedia\footnote{\href{https://en.wikipedia.org/wiki/Wikipedia:Redirect}{https://en.wikipedia.org/wiki/Wikipedia:Redirect}}. 
Using this method, we were able to Wikify a corpus of nearly 10 billion sentences with satisfactory precision, providing highly valuable results for various downstream NLP applications.\footnote{To keep the anonymity of this submission, more details will be shared in the final version.}
RedW aims to wikify as many mentions as possible, not only named entities \cite{AIDA,ACE-MSNBC} or salient terms \cite{ERD'14,AQUAINT}. 
We further present an efficient method for scoring RedW results, allowing a user to identify a subset of terms over which to apply heavier, more precision oriented methods. We demonstrate that the potential performance gain from using this method is substantial. This combined approach enables tuning to a desired point in the run-time - precision space. 



\section{Related work}
\label{sec:related}
A straightforward approach to context-free EL is selecting the most likely Wikipedia title of a given term, based on anchor dictionaries such as cross-wikis \citep{L12-1109}. This has been shown to be a strong baseline for EL \citep{chang, ACE-MSNBC,Fader09scalingwikipedia-based}. Here, instead of relying on static dictionaries generated from crawling the web \citep{L12-1109} or by aggregating anchor links in Wikipedia, we leverage the dynamic and powerful machinery of Wikipedia redirects.
Redirects have been incorporated before in e2e Wikification systems as part of the Spotting step \citep{TAGME, WAT, wikify}. 
However, RedW uses redirects to further solve the EL task; hence, to the best of our knowledge, it is the first context-free e2e Wikification system, 

Existing e2e Wikification methods ~\cite{WAT,TAGME} use context dependent EL algorithms, and thus focus on performance rather than runtime. 
Moreover, the computational requirements of EL methods have increased dramatically 
with the flourishing of neural models,  \citep{conf/acl/HeLLZZW13,DBLP:journals/corr/YamadaS0T16,DBLP:journals/corr/GaneaH17,Sun:2015:MMC:2832415.2832435}
which have established state-of-the-art performance results, while further increasing computational demands.

In contrast, \cite{nguyen2014aida} 
consider an efficient two-stage mapping algorithm. It first identifies a set of “easy” mentions with low ambiguity and links them to entities in a highly efficient manner. Next, it harnesses the high-confidence linkage of “easy” mentions to establish more reliable contexts 
to disambiguate the remaining mentions.
However, this work focuses only on the NER problem, rather than on the full problem of spotting and linking. 
Another run-time oriented work is \cite{hogan2012scalable}; 
there, the authors focus on consolidating entity linking datasets by inter-linking the individual datasets (finding links between equivalent entities) rather than spotting and linking a corpus from scratch. 

%
\section{RedW: A Redirect-Based Wikification}
\label{sec:termWikifier}
The e2e Wikification task consists of (1) \textbf{Spotting} -- detecting terms that are candidates to become mentions; (2) \textbf{Entity Linking} -- discard (prune) candidate terms or link them to Wikipedia entities. 
RedW relies on the assumption that a term - up to some simple editorial transformations - often matches its Wikipedia title, or a corresponding  Wikipedia \textit{redirect} page.
Thus, RedW harnesses the wisdom of the crowd as expressed in ``redirect'' pages, to find the terms to be Wikified, and link them to the appropriate Wikipedia article.
A main advantage of using redirects over anchor dictionaries ~\cite{L12-1109} is their dynamic nature. They are maintained and updated both automatically and by Wikipedia editors \cite{hill2014consider}, hence are expected to contain less noise compared to 
automatically created dictionaries. 
 


\subsection{Spotting}
\label{sec:spotting}
We pre-process Wikipedia\footnote{We use the dump enwiki-20180901 from Sep 2018} and build \textit{spotMap} -- a table of all Wikipedia titles, including the redirect titles. 
Given a new text, we try to match its n-grams to \textit{spotMap}, preferring  longer matches. Thus, we start with $n=10$, and gradually move to smaller $n$'s. 
Per n-gram, we check whether it appears in \textit{spotMap}, after capitalizing its first letter. 
Matched n-grams are kept as candidate mentions for the next step. Further, their tokens are marked to avoid including them in subsequent -- shorter -- candidate n-grams. 
The final list of matched n-grams is analyzed by 
the EL step.

\subsection{Entity Linking}
\label{sec:base-TW}
Given a candidate term and its matched title/redirect, 
RedW aims to link it to the corresponding Wikipedia page, unless
the resulting page is a disambiguation page
\footnote{I.e., has the text \textit{may [also] refer to} in its first sentence}.
Our hypothesis is that having two evidences - both a term and a matching title/redirect - increases the likelihood that this is a correct mention.
Discarding terms corresponding to disambiguation pages reduces recall. 
We therefore propose 
RedW+, which when given a term and a matching disambiguation page, links the term to its most common page among all pages pointed from the disambiguation page. 
Clearly, the increase in recall comes at the expense of precision. As mitigation, we demonstrate how our suggested confidence score can be used to increase precision.

%


\section{Identifying potential system errors}
\label{sec:RedWerrodDetection}
Our next step is to provide an efficient method to properly estimate the confidence of RedW results, such that RedW errors will be characterized by low-confidence scores, which will be further Wikified by heavier tools.
To this end, we present a corresponding large-scale error-analysis of RedW results over Wikipedia.
There are two types of error:
i) linking a term to an incorrect page, while missing the correct page; ii) linking a term to a page while there is no Wikipedia page that corresponds to the meaning of the term in the sentence where it was spotted. 
We leverage Wikipedia internal links (WL) as ground truth against which we compare RedW results. 
Given an article in which RedW resolves term $s$ to a Wikipedia page $C$, we assume that RedW is correct if the article contains a WL to $C$. The agreement between RedW and WL is defined on the article level since in Wikipedia
there is at most one WL per concept in an article, usually but not necessarily in the first time it is referred to. 
Moreover, this definition reduces noise originating from the noisy nature of WLs, which sometimes associate the concept to neighboring terms instead of to the concept term itself. 
  
We treat articles where RedW detected a mention $s$ as independent Bernoulli variables, which take a value $1$ if RedW agrees with WL and $0$ otherwise.  
Let $N$ be the number of articles where RedW detects $s$ as a mention, and let $K$ be the number of articles among $N$ where RedW agrees with WL. The empirical probability of agreement between RedW and WL is $K/N$. We denote this probability by $SR$ (Success Rate). Notice that WLs provide only a partial coverage of the references to most concepts in Wikiepdia. An article can refer to a concept $C$ but have no WL to $C$, while in other cases the absence of WL is justified, namely the term does not refer to any concept. Thus, a disagreement between RedW and WL does not necessarily mean that RedW is incorrect. Since $SR$ measures the \textit{agreement} between RedW and WL, it underestimates the true probability for a correct answer by RedW.
Since all terms ``suffer" from the partial coverage of WLs, ranking the terms by $SR$ could have served our objective of sorting RedW results. However, different concepts have different levels of coverage by WLs, an effect that we would like to neutralize. 
Thus, we normalize $SR$ by its highest value among all the terms that are mapped to the same concept by RedW.  We call this normalized score $SR_{norm}$. 
The assumption behind this normalization is that for most concepts, among all the terms which may refer to it, there is at least one term which does not have multiple senses. The $SR$ of this term is expected to suffer only from the partial coverage problem of that concept, thus normalizing all other terms of the concept by this value should neutralize this effect. 
In Section \ref{sec:experiments} we show that $SR_{norm}$ significantly outperforms other scores in predicting RedW/RedW+ errors.

\section{Experiments}
\label{sec:experiments}
\subsection{Performance of RedW}
We first evaluate the performance of RedW compared to other e2e Wikification methods. Note, that RedW 
aims to Wikify 
all terms in a text which can be linked to a Wikipedia page, independent of their type and role. Thus entity linking datasets which focus either on NE ~\cite{AIDA,ACE-MSNBC} or on salient terms ~\cite{AQUAINT} are sub-optimal for this evaluation. 
We chose two comprehensive Wikification datasets for evaluation: Wiki-test and Trans-test (a dataset of spoken data) from ~\cite{YosiBenchmark}, and two datasets, IITB ~\cite{IITB} and Aquiant~\cite{AQUAINT}, that contain both named-entities and other entity types, but are not comprehensive as the former two. We compare RedW to two e2e widely used solutions: Tagme \cite{TAGME} and WAT \cite{WAT}\footnote{Using the best configuration for detecting all entity types}. 
Further, we employed GERBIL\footnote{http://gerbil.aksw.org/gerbil/} through which we also compared experimental results against AIDA~\cite{AIDA}, DBpedia Spotlight\footnote{https://www.dbpedia-spotlight.org/}~\cite{DBPediaSL}, 
and Babelfy~\cite{Babelfy}. 
Since IITB and Aquaint do not, in practice, include all mentions, we add IITB* and Aquaint* which are the result when using the method of evaluation described in \cite{EREL}, where an annotation is considered as false-positive if its term is matched or overlapped with a term of the benchmark, but the linked concept is different. Table \ref{tab:accuracy} shows that (1) RedW achieves the best F1 on all datasets except Aquaint*;
(2) RedW, TagMe and WAT, which support all types of entities achieve high recall on all datasets, while the systems focusing on named entities only, achieve very low recall; (3) RedW achieves the highest precision on all datasets 
while maintaining a relatively high recall; (4) RedW+ increases the recall at the cost of precision; 
This is expected as disambiguation pages are more noisy and the simple commonness is a relatively naive solution. 
We discuss in the next section what a heavier method could achieve on the hard mentions.    

\begin{table*}[h!]
\small
	\begin{center}
    \small
		\begin{tabular}{l|c|c|c||c|c|c||c|c|c||c|c|c||c|c|c|} 
			\cline{2-16}
			\multirow{2}{*}{} & 
				\multicolumn{3}{c||}{\textbf{IITB}} &
				\multicolumn{3}{c||}{\textbf{IITB*}} &
				\multicolumn{3}{c||}{\textbf{Trans}} &
				\multicolumn{3}{c||}{\textbf{Wiki}} &
				\multicolumn{3}{c||}{\textbf{Aquaint*}} \\
			\cline{2-16}
			 & P & R & F1  & P & R & F1 & P & R & F1 & P & R & F1 & P & R & F1 \\ 
			 \hline
			 \multicolumn{1}{|l|}{TagMe} &
			  .12  & .48 & .19 & .65 & .48 & .55 & .33 & .51 & .41 & .42 & .58 & .49 & .81  & \textbf{.89} & \textbf{.85}\\
			\hline 
			 \multicolumn{1}{|l|}{WAT} &
			 .18 & .53 & .27 & .66 & .53 & .59 & .44 & .39 & .42 & .44 & .45 & .45 & .85 & .84 & .84 \\
            \hline 
             \multicolumn{1}{|l|}{AIDA} &
            \textbf{.51} & .16 & .20 & .62 &.15 & .19 & .05 & .01 & .02 & .27 & .08 & .14 & \textbf{.90} & .36 & .52  \\ 
            \hline 
             \multicolumn{1}{|l|}{SpotLight} & 
            .51 & .26 & .35  & .67 & .25 & .36 & .33 & .09 & .15 & .59 & .21 & .31 &  .82 & .46 & .59 \\ 
            \hline 
             \multicolumn{1}{|l|}{Babelfy} & 
           .22 & .17 & .18  & .41 & .17 & .20 & .03 & .01 & .02 & .22 & .07 & .12 & .76 & .36 & .50  \\ 
            \hline
			\multicolumn{1}{|l|}{RedW} &
			 .26 & .67 & \textbf{.37} & .78 & .69 & .73 & \textbf{.76} & .62 & \textbf{.68} & \textbf{.72} & .64 & \textbf{.68} & .86 & .78 & .82 \\
            \hline 
			\multicolumn{1}{|l|}{RedW+} & 
			.19 & \textbf{.69} & .30 & \textbf{.80} & \textbf{.69} & \textbf{.74} & .57 & \textbf{.64} & .60 & .57 & \textbf{.67} & .62 & .83 & .79 & .81 \\
			\hline
		\end{tabular}
		\caption{\small Comparison of e2e methods - $Precision, Recall, F1$.}
         \label{tab:accuracy}       
       \end{center}
       \label{RedWResults}
\end{table*}

\subsection{Runtime performance}
Table~\ref{tab:runtime} reports runtime of RedW and TagMe on short text snippets of size 100-600 characters, as well as on a large news corpus of size 600Gb.   
The experiments were done on a single Linux-based computer comprising 14 processors in each of 2 cores and a total memory of 60 Gb.
RedW runtime is improved by more than an order of magnitude compared to TagMe.

\begin{table}[tbh]
\small
	\begin{center}
    \small
		\begin{tabular}{|l|c|c|} 
		    \hline 
		        & Short snippets &  Large corpus \\
			\hline
			 TagMe & 113.96 ms & 22 days \\
			\hline
			  RedW & 5.3 ms & 41 hours \\
			\hline  
		\end{tabular}
		\caption{\small Runtime performance}
         \label{tab:runtime}       
       \end{center}
\end{table}

\subsection{Confidence score}
\begin{figure}[ht!]
\centering
\includegraphics[width=75mm]{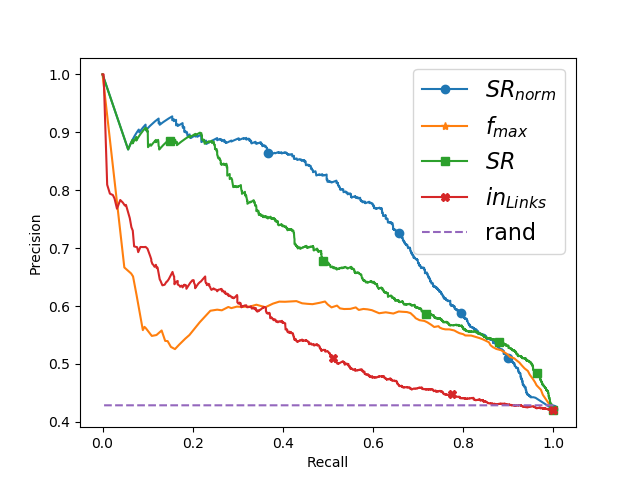}
\caption{\small Precision (fraction of errors in the subset) vs. Recall (fraction of detected errors) on Trans-test. 
\label{fig:RedErrors}}
\end{figure}

We compare $SR_{norm}$ (see Section \ref{sec:RedWerrodDetection}) to several baselines: (1) $rand$ -- the trivial random score; (2) $f_{max}$ -- the frequency of $C$ among concepts that $s$ is linked to, computed based on WLs population; (3) ${in}_{links}$ -- the number of incoming WLs to $C$. Note that ${in}_{links}$ assigns the same score to all the terms that are linked to $C$ by RedW; (4) $SR$ -- to examine the effectiveness of the intra-concept normalization performed in $SR_{norm}$.
Clearly, the potential precision gain from applying heavy Wikification methods increases with the error rate within the subset on which they are applied. We therefore evaluate the methods on the task of detecting errors by RedW+, which provides higher recall and lower precision compared to RedW. 
Figure \ref{fig:RedErrors} shows precision-recall curves for the task of detecting RedW+ errors, when different scoring methods are used to rank the elements in the Wiki-test dataset. Clearly, $SR_{norm}$ significantly outperforms all other scores. 

\begin{figure}[ht!]
\centering
\includegraphics[width=65mm]{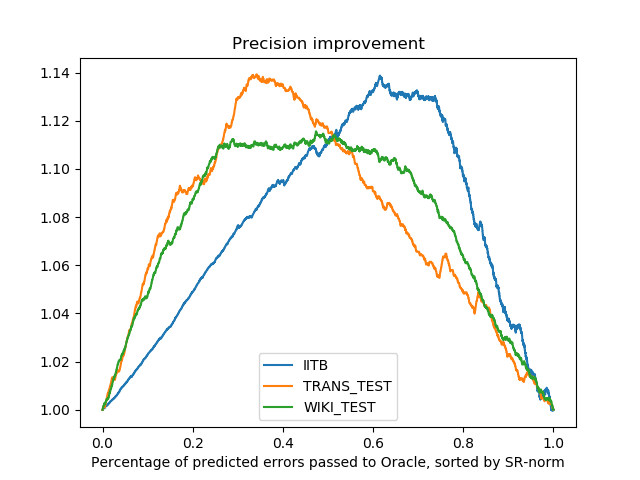}
\caption{Ratio between precision obtained by an ideal system as a result of selecting the subset using $SR_{norm}$ and a random selection of the subset.  \label{fig:improvement}}
\end{figure}

Next we examine the potential performance gain of using $SR_{norm}$ for selecting the subset of texts to be processed by heavy methods. To this end we assume an ideal method, able to correct all the disambiguation errors of the system and prune all terms mistakenly detected as mentions. Figure~\ref{fig:improvement} shows the ratio of improvement in the precision of RedW+, when the subset is selected by $SR_{norm}$, 
compared to selecting at random. 
The results demonstrate the potential effectiveness of the $SR_{norm}$ scoring method. The maximal gain exceeds $10\%$ in all datasets. In the region of interest (small subsets), the improvement increases monotonically.


\vspace{-10pt}
\section{Conclusions and future work}
\label{sec:conclusions}
We present a new, run-time oriented 
tool for context-free Wikification based on Wikipedia redirects, which allows to Wikify large corpora. Despite its simplicity, its performance is superior to other, heavier e2e systems on several datasets. We 
further show how to properly estimate the confidence of the system predictions,
and demonstrate the associated potential performance gain. 
The $SR_{norm}$ score can also be used to score entire documents (e.g., by averaging over all annotations in the document), 
highlighting documents over which combining RedW with heavy \textit{global} EL algorithms would be most promising.

\bibliography{emnlp-TW}

\begin{thebibliography}{24}
\expandafter\ifx\csname natexlab\endcsname\relax\def\natexlab#1{#1}\fi

\bibitem[{Berant et~al.(2013)Berant, Chou, Frostig, and
  Liang}]{conf/emnlp/BerantCFL13}
Jonathan Berant, Andrew Chou, Roy Frostig, and Percy Liang. 2013.
\newblock Semantic parsing on freebase from question-answer pairs.
\newblock In \emph{EMNLP}, pages 1533--1544. ACL.

\bibitem[{Carmel et~al.(2014)Carmel, Chang, Gabrilovich, Hsu, and
  Wang}]{ERD'14}
David Carmel, Ming-Wei Chang, Evgeniy Gabrilovich, Bo-June~(Paul) Hsu, and
  Kuansan Wang. 2014.
\newblock Erd 14: Entity recognition and disambiguation challenge.
\newblock In \emph{ACM SIGIR Forum}, volume~48.

\bibitem[{Chang et~al.(2016)Chang, Spitkovsky, Manning, and Agirre}]{chang}
Angel~X. Chang, Valentin~I. Spitkovsky, Christopher~D. Manning, and Eneko
  Agirre. 2016.
\newblock Evaluating the word-expert approach for named-entity disambiguation.
\newblock \emph{CoRR}, abs/1603.04767.

\bibitem[{Fader et~al.(2009)Fader, Soderland, and
  Etzioni}]{Fader09scalingwikipedia-based}
Anthony Fader, Stephen Soderland, and Oren Etzioni. 2009.
\newblock Scaling wikipedia-based named entity disambiguation to arbitrary web
  text.
\newblock In \emph{In Proceedings of the IJCAI Workshop on User-contributed
  Knowledge and Artificial Intelligence: An Evolving Synergy}.

\bibitem[{Ferragina and Scaiella(2012)}]{TAGME}
Paolo Ferragina and Ugo Scaiella. 2012.
\newblock Fast and accurate annotation of short texts with wikipedia pages.
\newblock \emph{IEEE Software 29(1)}.

\bibitem[{Ganea and Hofmann(2017)}]{DBLP:journals/corr/GaneaH17}
Octavian{-}Eugen Ganea and Thomas Hofmann. 2017.
\newblock Deep joint entity disambiguation with local neural attention.
\newblock \emph{CoRR}, abs/1704.04920.

\bibitem[{He et~al.(2013)He, Liu, Li, Zhou, Zhang, and
  Wang}]{conf/acl/HeLLZZW13}
Zhengyan He, Shujie Liu, Mu~Li, Ming Zhou, Longkai Zhang, and Houfeng Wang.
  2013.
\newblock Learning entity representation for entity disambiguation.
\newblock In \emph{ACL (2)}, pages 30--34. The Association for Computer
  Linguistics.

\bibitem[{Hill and Shaw(2014)}]{hill2014consider}
Benjamin~Mako Hill and Aaron Shaw. 2014.
\newblock Consider the redirect: A missing dimension of wikipedia research.
\newblock In \emph{Proceedings of The International Symposium on Open
  Collaboration}, page~28. ACM.

\bibitem[{Hoffart et~al.(2011)Hoffart, Yosef, Bordino, Furstenau, Pinkal,
  Spaniol, Taneva, Thater, and Weikum}]{AIDA}
Johannes Hoffart, Mohamed~A. Yosef, Ilaria Bordino, Hagen Furstenau, Manfred
  Pinkal, Marc Spaniol, Bilyana Taneva, Stefan Thater, and Gerhard Weikum.
  2011.
\newblock Robust disambiguation of named entities in text.
\newblock In \emph{Proceedings of the Conference on Empirical Methods in
  Natural Language Processing}, pages 782--792.

\bibitem[{Hogan et~al.(2012)Hogan, Zimmermann, Umbrich, Polleres, and
  Decker}]{hogan2012scalable}
Aidan Hogan, Antoine Zimmermann, J{\"u}rgen Umbrich, Axel Polleres, and Stefan
  Decker. 2012.
\newblock Scalable and distributed methods for entity matching, consolidation
  and disambiguation over linked data corpora.
\newblock \emph{Web Semantics: Science, Services and Agents on the World Wide
  Web}, 10:76--110.

\bibitem[{Kulkarni et~al.(2009)Kulkarni, Singh, Ramakrishnan, and
  Chakrabarti}]{IITB}
Sayali Kulkarni, Amit Singh, Ganesh Ramakrishnan, and Soumen Chakrabarti. 2009.
\newblock Collective annotation of wikipedia entities in web text.
\newblock In \emph{KDD}.

\bibitem[{Levy et~al.(2017)Levy, Gretz, Sznajder, Hummel, Aharonov, and
  Slonim}]{W17-5110}
Ran Levy, Shai Gretz, Benjamin Sznajder, Shay Hummel, Ranit Aharonov, and Noam
  Slonim. 2017.
\newblock Unsupervised corpus--wide claim detection.
\newblock In \emph{Proceedings of the 4th Workshop on Argument Mining}, pages
  79--84. Association for Computational Linguistics.

\bibitem[{Mass et~al.(2018)Mass, Kotlerman, Mirkin, Venezian, Witzling, and
  Slonim}]{YosiBenchmark}
Yosi Mass, Lili Kotlerman, Shachar Mirkin, Elad Venezian, Gera Witzling, and
  Noam Slonim. 2018.
\newblock What did you mention? a large scale mention detection benchmark for
  spoken and written text.
\newblock \emph{CoRR}, abs/1801.07507.

\bibitem[{Mendes et~al.(2011)Mendes, Jakob, Garc\'{\i}a-Silva, and
  Bizer}]{DBPediaSL}
Pablo~N. Mendes, Max Jakob, Andr{\'e}s Garc\'{\i}a-Silva, and Christian Bizer.
  2011.
\newblock Dbpedia spotlight: Shedding light on the web of documents.
\newblock In \emph{Proceedings of the 7th International Conference on Semantic
  Systems}, I-Semantics '11, pages 1--8, New York, NY, USA. ACM.

\bibitem[{Mihalcea and Csomai(2007)}]{wikify}
Rada Mihalcea and Andras Csomai. 2007.
\newblock Wikify!: Linking documents to encyclopedic knowledge.
\newblock In \emph{CIKM}, pages 233--242, New York, NY, USA. ACM.

\bibitem[{Milne and Witten(2008)}]{AQUAINT}
David Milne and Ian~H. Witten. 2008.
\newblock Learning to link with wikipedia.
\newblock In \emph{CIKM}, pages 509--518.

\bibitem[{Moro et~al.(2014)Moro, Raganato, and Navigli}]{Babelfy}
Andrea Moro, Alessandro Raganato, and Roberto Navigli. 2014.
\newblock Entity linking meets word sense disambiguation: a unified approach.
\newblock \emph{Transactions of the Association for Computational Linguistics},
  2:231--244.

\bibitem[{Nguyen and Trong~Hai(2017)}]{EREL}
Cuong Nguyen and Duong Trong~Hai. 2017.
\newblock Erel: an entity recognition and linking algorithm.
\newblock \emph{Journal of Information and Telecommunication}, 2:1--20.

\bibitem[{Nguyen et~al.(2014)Nguyen, Hoffart, Theobald, and
  Weikum}]{nguyen2014aida}
Dat~Ba Nguyen, Johannes Hoffart, Martin Theobald, and Gerhard Weikum. 2014.
\newblock Aida-light: High-throughput named-entity disambiguation.
\newblock \emph{LDOW}, 1184.

\bibitem[{Piccinno and Ferragina(2014)}]{WAT}
Francesco Piccinno and Paolo Ferragina. 2014.
\newblock From tagme to wat: a new entity annotator.
\newblock In \emph{ERD, the first international workshop on Entity recognition
  \& disambiguation}, pages 55--62.

\bibitem[{Ratinov et~al.(2011)Ratinov, Roth, Downey, and Anderson}]{ACE-MSNBC}
Lev-Arie Ratinov, Dan Roth, Doug Downey, and Mike Anderson. 2011.
\newblock Local and global algorithms for disambiguation to wikipedia.
\newblock In \emph{Proceedings of the 49th Annual Meeting of the Association
  for Computational Linguistics}, pages 1375--1384.

\bibitem[{Spitkovsky and Chang(2012)}]{L12-1109}
Valentin~I. Spitkovsky and Angel~X. Chang. 2012.
\newblock A cross-lingual dictionary for english wikipedia concepts.
\newblock In \emph{Proceedings of the Eighth International Conference on
  Language Resources and Evaluation (LREC-2012)}. European Language Resources
  Association (ELRA).

\bibitem[{Sun et~al.(2015)Sun, Lin, Tang, Yang, Ji, and
  Wang}]{Sun:2015:MMC:2832415.2832435}
Yaming Sun, Lei Lin, Duyu Tang, Nan Yang, Zhenzhou Ji, and Xiaolong Wang. 2015.
\newblock Modeling mention, context and entity with neural networks for entity
  disambiguation.
\newblock In \emph{Proceedings of the 24th International Conference on
  Artificial Intelligence}, IJCAI'15, pages 1333--1339. AAAI Press.

\bibitem[{Yamada et~al.(2016)Yamada, Shindo, Takeda, and
  Takefuji}]{DBLP:journals/corr/YamadaS0T16}
Ikuya Yamada, Hiroyuki Shindo, Hideaki Takeda, and Yoshiyasu Takefuji. 2016.
\newblock Joint learning of the embedding of words and entities for named
  entity disambiguation.
\newblock \emph{CoRR}, abs/1601.01343.

\end{thebibliography}
\bibliographystyle{acl_natbib}

\end{document}